\documentclass[
  10pt,
  logo,
  onecolumn,
  copyright,
  nodate,
  colorlinks=true,
  allcolors=blue
]{nvidiatechreport}

\usepackage[numbers,sort,square]{natbib}
\usepackage{flafter}
\usepackage{booktabs}
\renewcommand*{\backref}[1]{}
\renewcommand*{\backrefalt}[4]{}
\renewcommand{\titlefont}{\color{nvidiagreen}\normalfont\bfseries\fontsize{15}{18}\selectfont}
\renewcommand{\headingfont}{\color{black}\bfseries\fontsize{12}{13}\selectfont}
\titlespacing*{\section}{0pc}{1.30625ex plus1.9pt minus1.425pt}{2.5pt}
\titlespacing*{\subsection}{0pc}{1.0925ex plus1.425pt minus0.95pt}{1pt}
\titlespacing*{\subsubsection}{0pc}{1ex plus1.25pt minus0.75pt}{1pt}

\title{Do Pretraining Optimizer Gains Transfer to KL Distillation?}
\author{\large Lavinia Ghita\textsuperscript{*}, Dhruv Desai\textsuperscript{*}\\
\normalfont NVIDIA}
\date{}

\begin{document}
\maketitle
\begingroup
\renewcommand{\thefootnote}{*}
\footnotetext{Equal contribution. Emails: \texttt{lghita@nvidia.com},
\texttt{dhruvd@nvidia.com}.}
\endgroup

\noindent\textbf{Abstract.}\enspace
Matrix-structured optimizers can improve language-model pretraining, but
it is unclear whether these gains transfer to short-horizon KL
distillation from an instruction-tuned initialization. We compare AdamW and
diagonal AdaGrad with two tractable Shampoo reductions, SOAP, and All-2D
Muon across four dense Qwen student--teacher pairs under a common
forward-KL objective and fixed training budget. Across scales, none of the
evaluated matrix-structured recipes consistently improves the
quality--trainer-time trade-off over the
elementwise references: AdamW generally attains the lowest validation KL,
while AdaGrad can be slightly faster at larger scale. SOAP often tracks
AdamW in KL but incurs substantial trainer-time overhead, and the two Shampoo
reductions reverse ordering across model pairs, showing that conclusions
depend on the feasible recipe. These results suggest that pretraining
optimizer gains do not transfer automatically to KL distillation and
should be re-established for the target objective and systems regime.
\par

\section{Introduction}

Prior work on Shampoo, SOAP, Muon, and related matrix-structured optimizers
reports data-efficiency or wall-clock gains over AdamW in language-model
pretraining~\cite{shampoo,soap,muon,muonpractical,transfer}. Those
comparisons concern long-horizon
next-token training from random initialization, with outcomes sensitive
to model scale, batch size, parameterization, and hyperparameter
transfer~\cite{distshampoo,muonpractical,transfer}. KL distillation instead
starts from an instruction-tuned student and uses the teacher's full next-token
probability distribution, rather than a one-hot target, at every
supervised token. It also runs for a comparatively short horizon, while
online teacher inference reduces the optimizer's share of trainer time.
Consequently, an optimizer advantage in pretraining does not necessarily
imply lower validation KL per training step or per unit of trainer time in
short-horizon distillation.

Within each student--teacher pair, all recipes share the tokenizer, data,
forward-KL objective, batch size, and training budget; we compare minimum
validation KL within budget, validation KL at matched steps, and trainer
time. Canonical two-sided Shampoo~\cite{shampoo} cannot
be applied uniformly to the evaluated matrix geometries under the
factor-dimension cap. We therefore derive a scalar-row-mean quarter-power
reduction and compare it with a right-only half-power reduction based on
Scalable Shampoo's one-sided construction~\cite{distshampoo}, treating
both as specific feasible recipes rather than family-wide representatives.

The rest of the paper is structured as follows:
Section~\ref{sec:methods} defines the distillation objective and optimizer
recipes, including the Shampoo reductions. Section~\ref{sec:experimental}
describes the experimental design, and Section~\ref{sec:results} reports
fixed-budget quality, optimization trajectories, and trainer time.
Section~\ref{sec:discussion} discusses interpretation and limitations,
and Section~\ref{sec:conclusion} concludes.
Appendix~\ref{app:implementations} documents implementation fidelity,
Appendix~\ref{app:filters} grafting, and Appendix~\ref{app:hp}
hyperparameter selection.

\section{Objective and optimizer recipes}
\label{sec:methods}

\subsection{KL distillation}
\label{sec:kl}

Knowledge distillation matches a student distribution $p_\theta$ to a
frozen teacher distribution $q$ by minimizing forward KL at temperature
one~\cite{hinton}. Let $q_{n,t}$ and $p_{\theta,n,t}$ denote their
distributions over the shared vocabulary at token position $t$ in example
$n$, and let $m_{n,t}\in\{0,1\}$ indicate a supervised assistant token. We
report the validation token mean
\begin{equation}
\mathcal{L}_{\mathrm{val}}(\theta)
=
\frac{1}{\sum_{n,t}m_{n,t}}
\sum_{n,t}m_{n,t}\,
D_{\mathrm{KL}}\!\left(q_{n,t}\,\|\,p_{\theta,n,t}\right).
\label{eq:kd}
\end{equation}
Validation weights tokens equally, so region contributions scale with
token count and longer reasoning regions generally dominate. A smaller
value indicates closer teacher matching. The blended training
loss~\cite{scaling} avoids this between-region length dependence by
weighting the answer and reasoning means separately within each example.
For example $n$, let $A_n$ and $C_n$ denote its nonempty sets of answer and
reasoning tokens, and let
$\ell_{n,t}=D_{\mathrm{KL}}(q_{n,t}\,\|\,p_{\theta,n,t})$. Its loss is
\begin{equation}
\mathcal{L}_n
=
\lambda\frac{1}{|A_n|}\sum_{t\in A_n}\ell_{n,t}
+(1-\lambda)\frac{1}{|C_n|}\sum_{t\in C_n}\ell_{n,t},
\label{eq:blend}
\end{equation}
and training averages $\mathcal{L}_n$ across examples. Thus, $\lambda$
weights the answer-region mean independently of region length. Prior
evaluation identified $\lambda=0.5$ as the most stable answer--reasoning
weighting~\cite{scaling}; we use it throughout. If either region is empty,
the other receives unit weight. The objective contains no hard-label
language-modeling term.

\subsection{Matrix-structured pretraining optimizers}

We use AdamW and diagonal AdaGrad as complementary elementwise references.
AdamW reflects standard language-model practice, while AdaGrad connects
directly to the structured preconditioners studied here; a different
distillation formulation on Nemotron reasoning traces also found slightly
higher MMLU-Pro with AdaGrad than with its best AdamW
configuration~\cite{teachersupport}. AdaGrad
accumulates a coordinatewise sum of squared gradients~\cite{adagrad}, while
AdamW decouples weight decay from Adam's adaptive update~\cite{adamw}. The
comparison includes three matrix-structured
optimizer families used in language-model pretraining. Across Shampoo,
SOAP, and Muon, structured updates act on two-dimensional parameters as
matrices rather than treating their entries as independent scalar
coordinates.

Shampoo~\cite{shampoo} constructs left- and right-side second-moment
accumulators for a matrix gradient and applies inverse matrix roots on both
sides, providing a tractable structured alternative to full-matrix
AdaGrad~\cite{adagrad}. Maintaining these accumulators and computing their
inverse roots adds cost. SOAP~\cite{soap} periodically
derives eigenbases from Shampoo-style left and right accumulators and runs
Adam in the resulting rotated coordinates; between basis refreshes, it
continues updating coordinatewise second moments.

Muon~\cite{jordanmuon} instead orthogonalizes a momentum matrix with a
Newton--Schulz iteration.
Its published recipe applies Muon to hidden-layer matrices and uses AdamW
for embeddings and the output head~\cite{jordanmuon}. Our
\emph{All-2D Muon} recipe instead applies the spectral update to every
two-dimensional parameter, including embeddings and the output head, and
uses Adam only for one-dimensional parameters. This assignment is part of
the evaluated recipe rather than an implementation detail. We choose it to
test the spectral update uniformly across matrix-valued student parameters,
without routing the embedding and output matrices through a separate
Adam-family optimizer. Published language-model studies report improved
data efficiency or lower
wall-clock cost for SOAP~\cite{soap} and
Muon~\cite{muon,muonpractical,transfer} under their respective pretraining
protocols.
KL-Shampoo and KL-SOAP~\cite{klshampoo} use KL divergence to derive
structured covariance estimators; that optimizer-internal construction is
distinct from the teacher--student output KL studied here, and we do not
evaluate those variants.

\subsection{Shampoo variants and matrix geometry}
\label{sec:shampoo}

Elementwise methods scale parameters independently. Full-matrix
AdaGrad~\cite{adagrad} captures correlations among all $mn$ entries of a
weight matrix but requires a preconditioner with $(mn)^2$ entries.
Shampoo~\cite{shampoo} instead maintains one factor per matrix axis. For
the gradient
$G_t\in\mathbb{R}^{m\times n}$ of a weight with $m$ row coordinates and
$n$ column coordinates, define
\begin{equation}
L_t=\epsilon I_m+\sum_{\tau=1}^{t}G_\tau G_\tau^{\mathsf T},
\qquad
R_t=\epsilon I_n+\sum_{\tau=1}^{t}G_\tau^{\mathsf T}G_\tau.
\label{eq:grams}
\end{equation}
$L_t\in\mathbb{R}^{m\times m}$ acts in row-coordinate space and
$R_t\in\mathbb{R}^{n\times n}$ acts in column-coordinate space. Storage falls
from $m^2n^2$ entries to $m^2+n^2$, while the full factors mix coordinates
and can change the gradient direction rather than merely rescale entries
independently. Here $\epsilon>0$ denotes regularization; the equations
suppress the additional numerical damping used by the inverse-root routine.

Canonical matrix Shampoo applies both factors:
\begin{equation}
\widetilde G_t=L_t^{-1/4}G_tR_t^{-1/4}.
\label{eq:quarter}
\end{equation}
For row-major vectorization $g_t=\operatorname{vec}_{r}(G_t)$,
Equation~\ref{eq:quarter} is equivalent to
\begin{equation}
\operatorname{vec}_{r}(\widetilde G_t)
=\left(L_t^{1/4}\otimes R_t^{1/4}\right)^{-1}g_t.
\label{eq:shampoo-vectorized}
\end{equation}
If every $G_\tau$ has rank at most $r$, the Scalable Shampoo
analysis~\cite{distshampoo} relates the full-matrix AdaGrad accumulator
$A_t=\epsilon I_{mn}+\sum_{\tau=1}^{t}g_\tau g_\tau^{\mathsf T}$ to these
factors through the positive-semidefinite bound
\begin{equation}
A_t\preceq r\,L_t^{1/2}\otimes R_t^{1/2}.
\label{eq:shampoo-bound}
\end{equation}
Thus, the two quarter powers form a rank-dependent structured
preconditioner motivated by a positive-semidefinite bound, not an exact
factorization of the full accumulator. If both factors grow proportionally
to $t$, they give the usual heuristic decay
$t^{-1/4}t^{-1/4}=t^{-1/2}$~\cite{shampoo}. Equation~\ref{eq:quarter} is
the reference point below.

Under our fixed factor-dimension cap of $8192$, the students' matrix
geometries~\cite{qwen3,qwen25} permit right factors for more parameters than
left factors. The embedding and output head have shape
$(\text{vocabulary},\text{hidden})$, making their left factor at least
$151{,}936$ square and their right factor at most $5120$ square. The MLP
gate and up projections have shape
$(\text{intermediate},\text{hidden})$, also placing the smaller hidden
dimension on the right; the down projection reverses these dimensions, and
every attention projection satisfies the cap. A right factor therefore
captures correlations in the hidden-feature or input-column space for
almost every two-dimensional weight without constructing vocabulary- or
intermediate-width matrix roots. We select right-sided preconditioning for
computational feasibility, not because column-space structure is assumed
to be intrinsically more informative. Because transposition exchanges
$L_t$ and $R_t$, the left/right label is layout-dependent. Here
\emph{right-only} refers to the checkpoint's PyTorch layout, in which
linear weights are stored as $(\text{output},\text{input})$ and embedding
weights as
$(\text{vocabulary},\text{hidden})$.

The first variant replaces the left matrix factor by a scalar
summary for every matrix-eligible two-dimensional weight, including
weights whose full left factor would fit. Define that scalar row mean as
\begin{equation}
s_t=\frac{\operatorname{tr}(L_t)}{m}
=\epsilon+\frac{1}{m}\sum_{\tau=1}^{t}\lVert G_\tau\rVert_{\mathrm F}^2.
\label{eq:scalar-stat}
\end{equation}
Replacing $L_t$ by $s_tI_m$ in Equation~\ref{eq:quarter} gives
\begin{equation}
\widetilde G_t=s_t^{-1/4}G_tR_t^{-1/4}.
\label{eq:scalar-quarter}
\end{equation}
We call this \emph{scalar-row-mean quarter-power Shampoo}. It preserves the
mean eigenvalue of $L_t$ and canonical Shampoo's quarter-power
structure~\cite{shampoo} but removes every directional distinction in
output-row space; it reproduces $L_t^{-1/4}$ as an operator only when
$L_t=s_tI_m$. It has a right matrix factor, but it is not a pure right-only
method.

The second variant follows the one-sided construction of Scalable
Shampoo~\cite{distshampoo}. For $p,q\in[1,\infty]$ satisfying
$1/p+1/q=1$, with
$1/\infty=0$, that construction uses
\begin{equation}
\widetilde G_t=L_t^{-1/(2p)}G_tR_t^{-1/(2q)}.
\label{eq:asymmetric-shampoo}
\end{equation}
The symmetric choice $p=q=2$ recovers
Equation~\ref{eq:quarter}. Taking $p\to\infty$, with
$q=p/(p-1)\to1$, reaches the endpoint $(p,q)=(\infty,1)$, removes the
left factor, and gives
\begin{equation}
\widetilde G_t=G_tR_t^{-1/2}.
\label{eq:right-half}
\end{equation}
The half power is therefore the endpoint of the same Kronecker-factored
full-matrix-AdaGrad surrogate, rather than an independent exponent choice.
Scalable Shampoo introduces these one-sided endpoints for large layers,
applies them to embedding, softmax, and projection layers, and proves an
online-convex regret bound under uniformly rank-bounded
gradients~\cite{distshampoo}.
Omitting $L_t$ discards directional information in output-row coordinate
space, so this \emph{right-only half-power Shampoo} is not merely a
lower-cost implementation of the canonical preconditioner.

Thus, scalar-row-mean quarter-power Shampoo retains an isotropic left-side
scale and a right quarter-power direction, whereas right-only half-power
Shampoo removes the left factor and strengthens the right exponent.

For ideal, unstabilized Frobenius-norm grafting to a donor direction
$D$~\cite{grafting}, write $Q=G_tR_t^{-1/4}\neq0$ and
$P=s_t^{-1/4}Q$. Then
\begin{equation}
P\frac{\lVert D\rVert_{\mathrm F}}{\lVert P\rVert_{\mathrm F}}
=Q\frac{\lVert D\rVert_{\mathrm F}}{\lVert Q\rVert_{\mathrm F}},
\label{eq:scalar-graft}
\end{equation}
so, up to the implementation's $10^{-16}$ denominator stabilizer, the
scalar factor cancels and no scalar-row-mean information survives. We
therefore exclude this grafted variant as equivalent to the corresponding
normalized right-quarter direction. For right-only half-power Shampoo,
normalization preserves the non-scalar $R_t^{-1/2}$ direction while
replacing its norm with the donor norm, so the grafted and ungrafted
recipes remain distinct. The main comparison uses the ungrafted update;
Appendix~\ref{app:filters} discusses its grafted variant.

With blocking disabled, either evaluated recipe forms a right matrix factor
only when its dimension is at most $8192$. At $0.6\mathrm{B}$, every
two-dimensional parameter satisfies this cap and receives the right factor.
At $4\mathrm{B}$, $14\mathrm{B}$, and $32\mathrm{B}$, the MLP down
projection has shape $(\text{hidden},\text{intermediate})$, so its right
factor exceeds the cap and the parameter is instead updated with diagonal
AdaGrad. Embeddings, the output head, attention, and the gate and up
projections remain right-matrix-preconditioned at every scale;
one-dimensional parameters always use the diagonal update. Both recipes
use this right-factor/fallback rule; Scalable Shampoo instead combines
two-sided factors, selected one-sided factors, and blocking according to
layer geometry~\cite{distshampoo}.

With no momentum, cumulative statistics are updated before preconditioning
on every optimizer step. On the \emph{first preconditioner step}, the
resulting roots are used for that same update. Subsequent refreshes are
measured relative to this step, and the roots are held fixed between
refreshes. Before activation, matrix-preconditioned parameters use the raw
gradient; fallback parameters use the diagonal update throughout. The
learning rate and both schedule values are selected separately for each
method and student--teacher pair.
Appendix~\ref{app:hp} gives the selected values, and
Appendix~\ref{app:implementations} gives the numerical root procedure.

\section{Experimental design}
\label{sec:experimental}

\subsection{Models and data}

The first three pairs use the instruction-tuned Qwen3-0.6B, Qwen3-1.7B,
Qwen3-4B, Qwen3-8B, Qwen3-14B, and Qwen3-32B checkpoints~\cite{qwen3}:
$0.6\mathrm{B}\leftarrow1.7\mathrm{B}$,
$4\mathrm{B}\leftarrow8\mathrm{B}$,
$14\mathrm{B}\leftarrow32\mathrm{B}$. The fourth pair uses
Qwen2.5-32B-Instruct and Qwen2.5-72B-Instruct~\cite{qwen25}. We use Qwen2.5
for the largest pair because its released dense 32B and 72B checkpoints
have a student--teacher size ratio comparable to those of the Qwen3 pairs.
The Qwen3 checkpoints share a tokenizer and vocabulary; the two Qwen2.5
checkpoints likewise have byte-identical tokenizer artifacts. Although the
tokenizer and vocabulary differ between Qwen3 and Qwen2.5, the student and
teacher within each pair receive identical token IDs.

Each student is a dense decoder-only transformer. The hidden and MLP
intermediate sizes are respectively $(1024,3072)$, $(2560,9728)$,
$(5120,17408)$, and $(5120,27648)$ for the $0.6\mathrm{B}$,
$4\mathrm{B}$, $14\mathrm{B}$, and Qwen2.5 $32\mathrm{B}$ students. The
Qwen3 and Qwen2.5 vocabularies contain $151{,}936$ and $152{,}064$ entries,
respectively. The $0.6\mathrm{B}$ and $4\mathrm{B}$ checkpoints tie the
token embedding and output head; the $14\mathrm{B}$ and Qwen2.5
$32\mathrm{B}$ checkpoints use separate matrices of the same shape.
These dimensions determine the factor coverage described in
Section~\ref{sec:shampoo}.

The examples are chain-of-thought traces selected from NVIDIA's
Nemotron-Post-Training-Dataset-v2%
\footnote{\href{https://huggingface.co/datasets/nvidia/Nemotron-Post-Training-Dataset-v2}{Nemotron dataset card}}%
~\cite{nemotron,nemotronnano}. We retain
reasoning-on SFT traces with complete \texttt{<think>} and
\texttt{</think>} markup and create training and validation splits
stratified by the dataset's category field. The retained
\texttt{<think>} and \texttt{</think>} delimiters define the reasoning and
answer regions in Equation~\ref{eq:blend}. We apply a maximum sequence
length of $8192$.
Under both Qwen3 and Qwen2.5 tokenization, removal of overlength traces
leaves $79{,}535$ training examples and $1{,}006$ unique validation
examples.

\subsection{Comparison and evaluation protocol}
\label{sec:comparison}

We compare six distillation recipes: AdamW~\cite{adamw} and diagonal
AdaGrad~\cite{adagrad} as references; scalar-row-mean quarter-power and
right-only half-power Shampoo; SOAP~\cite{soap}; and All-2D
Muon~\cite{jordanmuon}. All-2D Muon applies its spectral update to every
two-dimensional parameter and Adam to one-dimensional parameters.

AdamW and AdaGrad use native PyTorch implementations; the remaining
methods use PyTorch/FSDP2 adaptations. SOAP matches the authors'
implementation in fixed-gradient parity tests. Trainer times are specific
to these adapters.

Every full-budget final configuration is evaluated with three seeds.
Qwen2.5 SOAP is limited to one matched 561-step run because of its measured
trainer-time overhead and is excluded from full-budget summaries.
Appendices~\ref{app:implementations}--\ref{app:hp} document implementation
details, grafting controls, selection, and final settings.

Within each student--teacher pair, every method uses the forward-KL
training loss in Equation~\ref{eq:blend}, with $\lambda=0.5$, and global
batch 144.
Weight decay is zero, no learning-rate scheduler or warmup is used, and
gradients are clipped to unit norm. Training uses bfloat16 FSDP2 with
float32 gradient reduction. Because weight decay is zero, the PyTorch
AdamW update is equivalent here to Adam; we retain the AdamW label.

All methods use the same fixed budget of 1683 optimizer updates. One
effective epoch is 552 updates, the number of complete batches of 144 in
the $79{,}535$-example training split, so the budget runs slightly past
three effective epochs. Pilot runs showed diminishing validation-KL
improvements toward the end of this horizon.

The three Qwen3 pairs use 72 NVIDIA GB200 accelerators across 18
four-accelerator nodes. The larger Qwen2.5
$32\mathrm{B}\leftarrow72\mathrm{B}$ pair uses 144 accelerators across 36
such nodes to accommodate the dense teacher and student. We use
\emph{trainer time} for elapsed time spent in training, validation, and
improvement-triggered checkpointing; it excludes model loading, queueing,
and final model export.

We evaluate quality using two criteria. The within-budget criterion is the
three-seed mean of each run's minimum validation KL observed during 1683
steps, corresponding to validation-based checkpoint selection. The
matched-step criterion compares the three-seed mean trajectories at common
evaluation steps. Absolute KL is compared only within a student--teacher
pair. We report means with one standard deviation and interpret both
quality views together with trainer time rather than combine them into a
single score. Comparisons are descriptive and use neither hypothesis tests
nor a fixed threshold for small KL differences. We describe an advantage
as consistent only when the same qualitative ordering recurs across model
pairs.

\section{Results}
\label{sec:results}

\subsection{Validation KL within a fixed budget}

\begin{figure}[ht]
\centering
\includegraphics[width=0.95\linewidth]{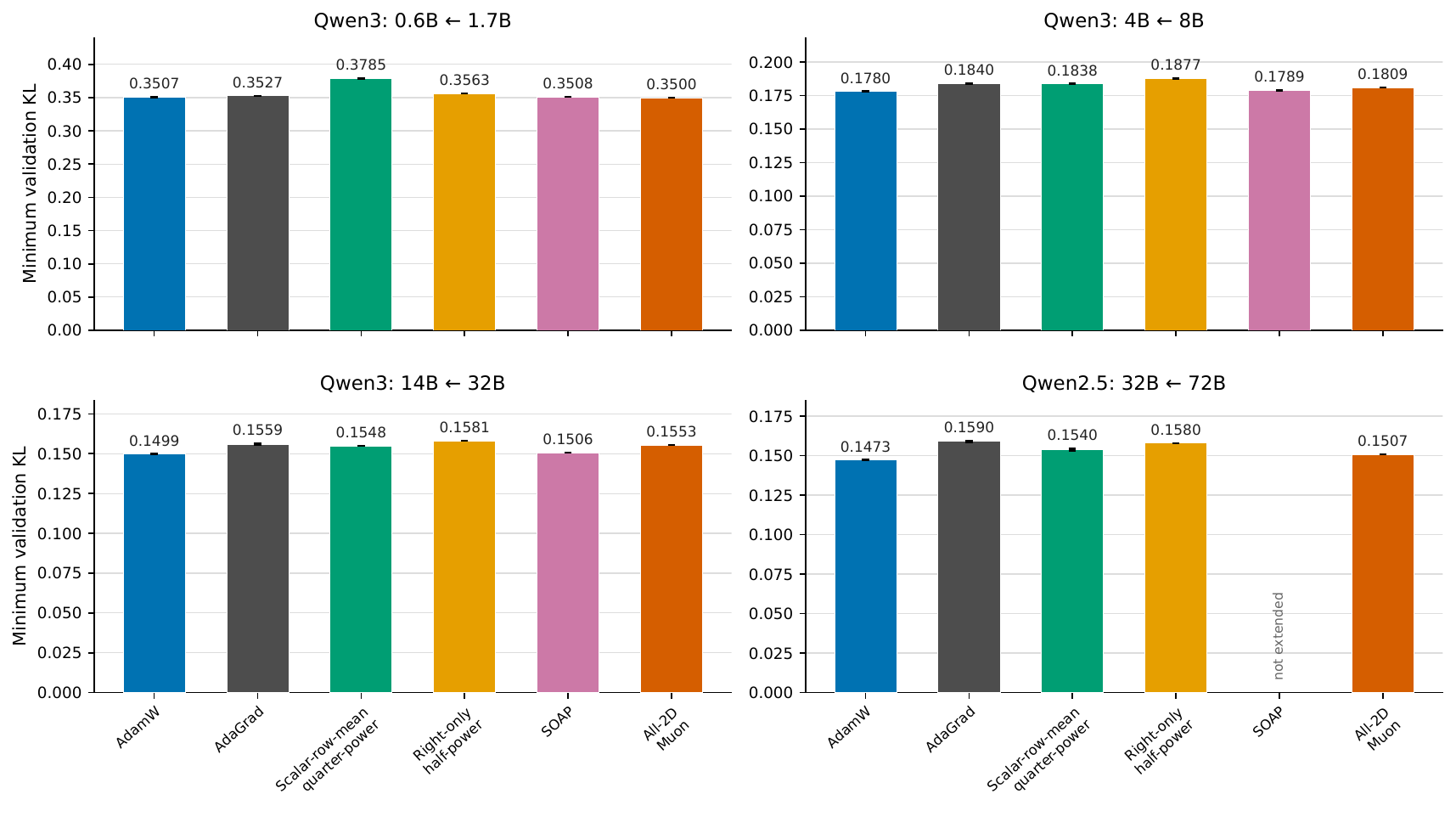}
\caption{Minimum validation KL within $1683$ steps: three-seed means of
per-seed minima, with one-standard-deviation error bars. Lower is better.
Qwen3 pairs use 72 GPUs and Qwen2.5 uses 144; Qwen2.5 SOAP was not
extended beyond one matched 561-step run.}
\label{fig:kl}
\end{figure}

Within the fixed budget, the evaluated matrix-structured methods show no
validation-KL advantage over AdamW that persists across model pairs
(Figure~\ref{fig:kl}).

AdamW, SOAP, and All-2D Muon are closely grouped at $0.6\mathrm{B}$;
AdamW is lower than SOAP and All-2D Muon at $4\mathrm{B}$ and
$14\mathrm{B}$, and lower than AdaGrad on all four pairs. The
Shampoo comparison is pair-dependent. Each Shampoo point is a
pair-specific, validation-selected configuration (Appendix~\ref{app:hp}),
not a family-wide optimum under an equal tuning budget. Right-only
half-power Shampoo has lower validation KL at $0.6\mathrm{B}$, whereas
scalar-row-mean quarter-power is lower at $4\mathrm{B}$; their nearly
identical trainer-time ratios indicate recipe dependence rather than a
quality--trainer-time trade-off. Scalar-row-mean quarter-power remains
above AdamW at $14\mathrm{B}$ and Qwen2.5 $32\mathrm{B}$.
The selected $k=10$ right-only half-power configuration is likewise above
AdamW at $14\mathrm{B}$, and the Qwen2.5 $32\mathrm{B}$ configuration is
above both AdamW and scalar-row-mean quarter-power.
Given that the teacher, student architecture, matrix geometry, and factor
coverage change together, these cross-pair differences do not establish a
scale trend.

\subsection{Optimization trajectories across model pairs}

\begin{figure}[ht]
\centering
\includegraphics[width=\linewidth]{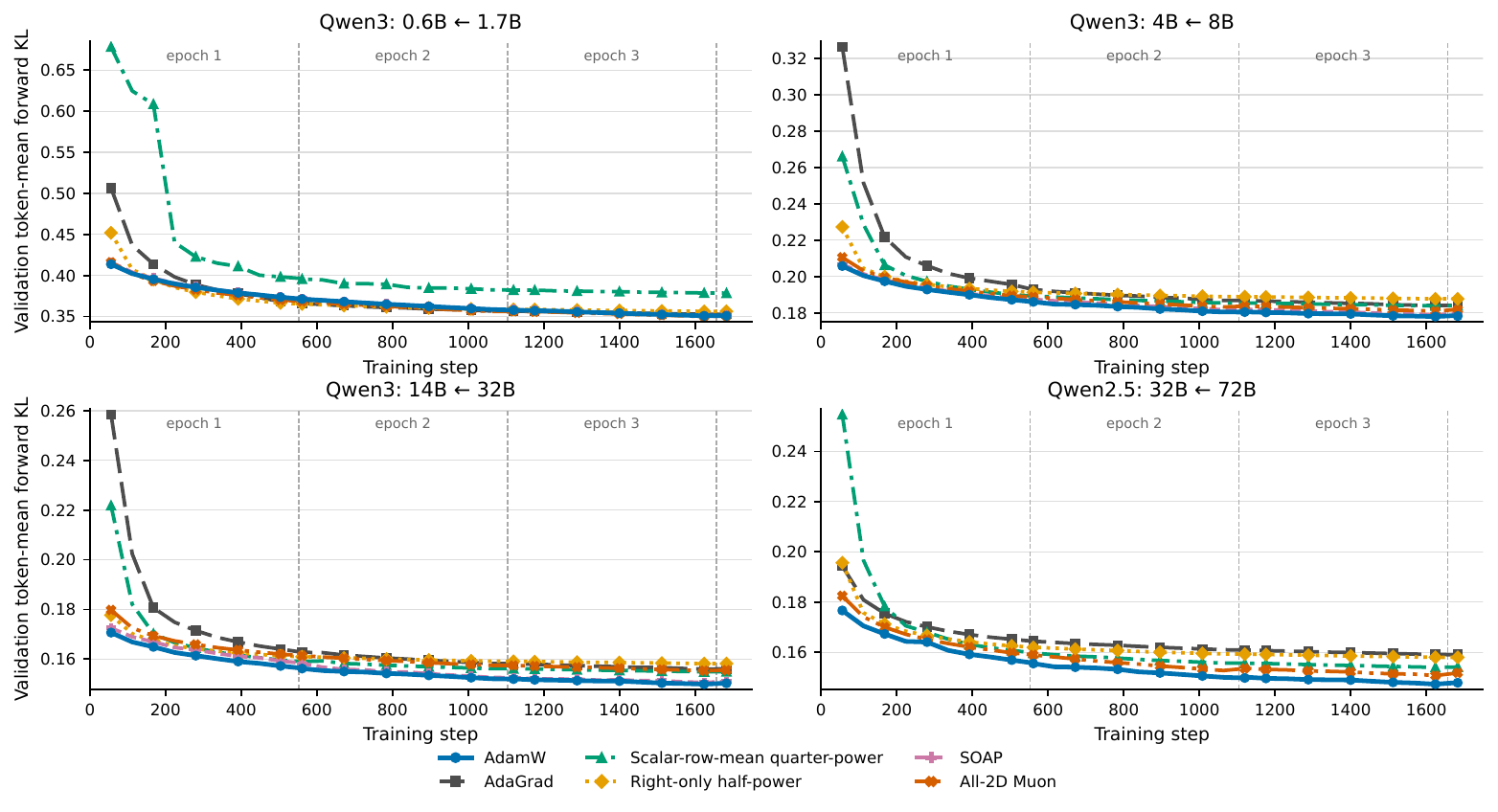}
\caption{Validation KL over training (three-seed means). Dashed lines mark
effective epochs of 552 updates; training ends at step $1683$. Vertical
scales differ.
Qwen3 pairs use 72 GPUs and Qwen2.5 uses 144; Qwen2.5 SOAP was not
extended beyond one matched 561-step run.}
\label{fig:curves}
\end{figure}

Matched-step trajectories show that optimizer ordering can depend on
training horizon, but no evaluated matrix-structured method shows a stable
early-step advantage across model pairs (Figure~\ref{fig:curves}). Across
methods, validation KL declines most steeply during the first effective
epoch and more slowly during the next two.

At $0.6\mathrm{B}$, AdaGrad and right-only half-power Shampoo are below
AdamW after roughly one effective epoch, but these advantages disappear by
the end of the budget. Scalar-row-mean quarter-power Shampoo remains above
AdamW at every recorded validation step; most of its initial gap closes
during the first effective epoch, but a positive gap remains at the final
update.

At $4\mathrm{B}$ and $14\mathrm{B}$, AdamW has lower validation KL than
every other completed method at each effective-epoch boundary. SOAP
approaches AdamW over training but does not overtake it at any recorded
step on those pairs. AdamW likewise remains lowest at each boundary on the
Qwen2.5 pair, where both Shampoo reductions remain above it. Thus, the
trajectories do not identify an advantage for matrix-structured methods
that is hidden by the fixed-budget summary.

\subsection{Trainer-time overhead}

\begin{figure}[ht]
\centering
\includegraphics[width=0.95\linewidth]{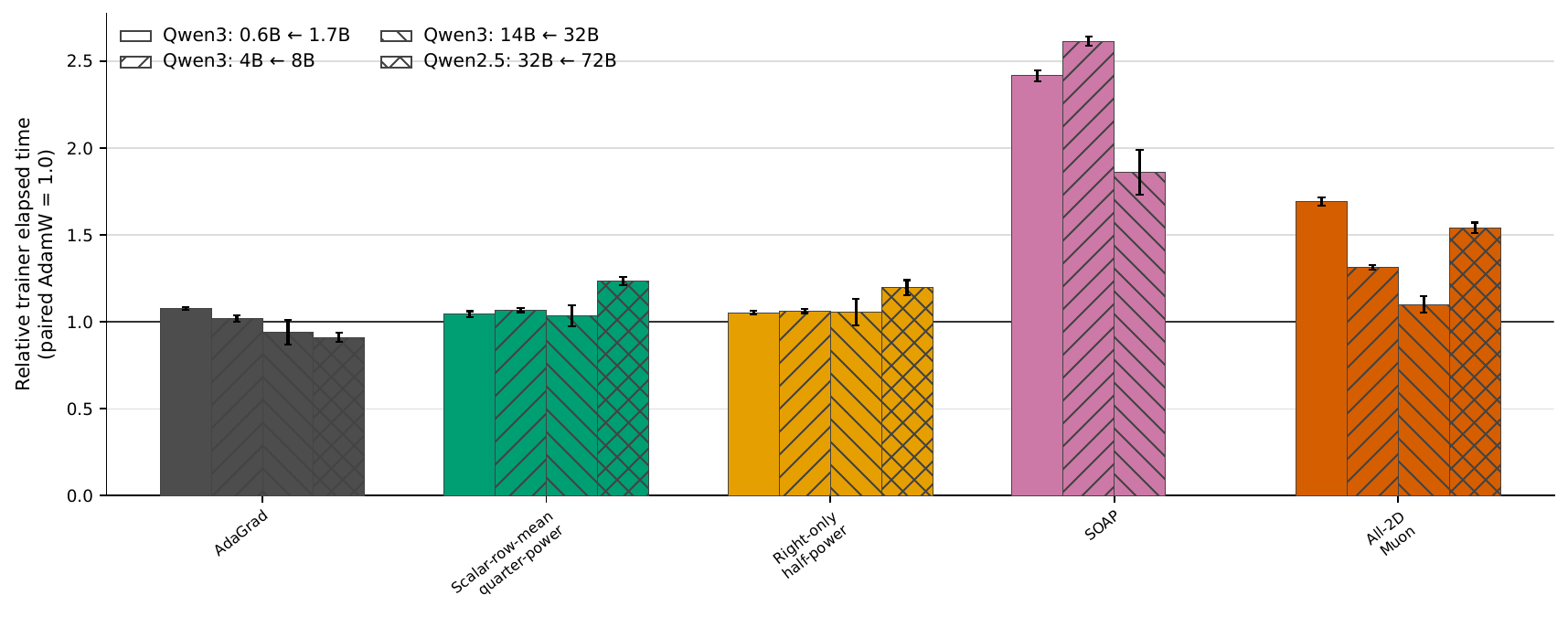}
\caption{Trainer time normalized per pair and seed to AdamW ($1.0$); bars
show three-seed means with one-standard-deviation error bars. Qwen3 pairs
use 72 GPUs and Qwen2.5 uses 144; ratios are within-pair. The matched
561-step Qwen2.5 SOAP run took $4.31\times$ AdamW trainer time and was not
extended.}
\label{fig:wall}
\end{figure}

Trainer-time results sharpen the quality comparison: every
matrix-structured method with completed results adds trainer time relative
to AdamW, whereas AdaGrad becomes slightly faster than AdamW at the larger
evaluated student sizes (Figure~\ref{fig:wall}). This is consistent
with AdaGrad's
lighter elementwise update: it maintains one accumulator rather than
AdamW's first- and second-moment states, reducing optimizer-state traffic
and pointwise work. The Qwen2.5 experiments use 144 GPUs, however, and
therefore do not provide a pure scale continuation of the fixed 72-GPU
regime used for Qwen3.

Among the evaluated matrix-structured methods, the two Shampoo reductions
have the smallest overhead and nearly identical trainer time on their
shared completed pairs. Their reversal in validation KL therefore reflects
the optimizer recipe rather than different compute budgets. All-2D Muon
becomes more expensive on the larger pairs. SOAP is qualitatively
different: it requires roughly two to three times AdamW's trainer time on
the Qwen3 pairs.

Online teacher inference, validation, and checkpointing reduce the
optimizer's share of total time. These ratios therefore measure
systems-level protocol overhead, not student-only optimizer-kernel
overhead, and may understate the latter.

At Qwen2.5 $32\mathrm{B}\leftarrow72\mathrm{B}$, SOAP incurred
substantially greater trainer time than AdamW without a clear improvement
in validation KL at the matched 561-step horizon. Its unblocked FSDP2
adapter gathers full gradients and moments, replicates matrix factors, and
refreshes bases frequently (Appendix~\ref{app:implementations}). Model
sharding does not distribute this optimizer work, so we did not extend
SOAP.

\section{Discussion and limitations}
\label{sec:discussion}

The transfer gap may reflect the shift from long-horizon pretraining to
short-horizon distillation from an instruction-tuned initialization.
Matrix-structured methods add estimation and refresh costs, leaving less
opportunity for quality gains to offset trainer-time overhead; the study
does not isolate these mechanisms causally.

That reversal comes from approximations that retain different information
and exponents, with factor coverage and selected hyperparameters that also
vary across pairs.

The SOAP and Muon conclusions are similarly tied to routing and
implementation choices. All-2D Muon differs from the hidden-layer-only
reference recipe, and unblocked SOAP differs algorithmically and
operationally from blocked distributed alternatives. Moreover, Shampoo
and SOAP use observed-gradient moments, which need not approximate the
model's Fisher information matrix~\cite{kunstner}. The findings therefore
do not rank optimizer
families independently of factorization, routing, and implementation.

\noindent\textbf{Limitations.}\enspace
The empirical scope is four dense Qwen student--teacher pairs from two
model generations and one chain-of-thought corpus. The study does not test
longer
schedules, other distillation objectives or data, or sparse model
architectures. Validation KL measures teacher matching rather than
downstream performance.

The evaluated Shampoo
recipes are unblocked and use at most one full right matrix factor.

Trainer times are specific to the 72- and 144-accelerator GB200
allocations and depend on the evaluated adapters. Qwen2.5 SOAP is
deliberately limited to one matched 561-step run and is excluded from
full-budget summaries. The study also includes no matched controls
initialized from checkpoints pretrained with each evaluated optimizer.
\par

\section{Conclusion}
\label{sec:conclusion}

Simple elementwise optimizers remain strong baselines for short-horizon
online KL distillation. Across four Qwen student--teacher pairs, AdamW
generally delivered the lowest validation KL, while AdaGrad provided
comparable efficiency and could be slightly faster for the larger LLMs
evaluated.
The evaluated matrix-structured methods did not produce a persistent
quality--trainer-time advantage, and the changing order of the two Shampoo
reductions showed that the exact feasible formulation matters. For
short-horizon online KL distillation, AdamW and AdaGrad should therefore be
the starting point, with matrix structure added only when it yields a
clear quality or trainer-time benefit in the target setting.

\vspace{2\baselineskip}

\clearpage
\appendix

\section{Optimizer implementations used}
\label{app:implementations}

AdamW and AdaGrad use \texttt{torch.optim}; the matrix-structured methods
use the PyTorch/FSDP2 implementations below. Method labels denote these
specific implementations, not entire optimizer families. Code for these
experiments will be released. Our main
\texttt{model-distillation-recipes} repository%
\footnote{\href{https://github.com/NVIDIA/model-distillation-recipes}{NVIDIA model-distillation-recipes repository}} documents the surrounding
distillation protocol, including data formatting and token masking.
Alternatives prefixed by ``Meta'' use the public Meta Distributed Shampoo
repository%
\footnote{\href{https://github.com/facebookresearch/optimizers/tree/d24a1491797e5bf2c2985d8789f3a28293d87a6f}{Meta Distributed Shampoo, revision \texttt{d24a149}}}%
~\cite{metashampoo}.
The implementation controls cover the three Qwen3 model pairs:
$0.6\mathrm{B}\leftarrow1.7\mathrm{B}$,
$4\mathrm{B}\leftarrow8\mathrm{B}$, and
$14\mathrm{B}\leftarrow32\mathrm{B}$. These controls identify faithful,
practical implementations rather than estimate scaling trends, so we do
not repeat them on the Qwen2.5 pair.

\subsection{Shampoo}
The two Shampoo recipes in the main comparison use our PyTorch/FSDP2
Scalable Shampoo backend. It ports selected equations and state-machine
behavior from the
Google Research implementation%
\footnote{\href{https://github.com/google-research/google-research/tree/1e55815f71e2cf4ee9c10c3d3fc2d8e2bf992ac4/scalable_shampoo}{Google Research Scalable Shampoo, revision \texttt{1e55815}}}, written in
JAX~\cite{jax} and released with Scalable Shampoo~\cite{distshampoo}. The
port retains epsilon-initialized Kronecker statistics, cumulative or
exponential-moving-average updates, separate statistic and inverse-root
schedules, delayed activation, SGD/AdaGrad norm grafting, separate donor
and Shampoo momentum, and the coupled-Newton inverse-root equations. It is
not a line-by-line translation: optimizer calls are indexed from one, only
two-dimensional parameters use Kronecker factors, and the
largest-eigenvalue estimate is deterministic across ranks.

The pinned Google implementation does not expose omission of an individual
Kronecker factor. Following the one-sided construction in the paper, we add
the right-only half-power update. The same backend also supports a
right-only quarter-power ablation and an AdaGrad-grafted right-only
half-power update, as well as scalar-sum and scalar-row-mean quarter-power
variants. Each scalar variant maintains one full-parameter scalar left
statistic even when the right factor is partitioned into blocks. For an
$m\times n$ gradient, the sum variant accumulates
$\lVert G_t\rVert_{\mathrm F}^2$ and the row-mean variant accumulates
$\lVert G_t\rVert_{\mathrm F}^2/m$. Non-matrix parameters and matrices whose active
factor exceeds the configured cap instead use diagonal AdaGrad and maintain
no scalar left statistic. The port further adds FSDP2 execution,
distributed checkpointing, and recipe fingerprints that reject
incompatible optimizer state. The pinned revision contains only the JAX
implementation. Its exponent override assigns one power to every Kronecker
factor, so the one-sided updates above are part of the port, as are the
scalar statistics and the distributed interfaces. A later public PyTorch
port of the same implementation%
\footnote{\href{https://github.com/google-research/google-research/tree/5b09c22d73a9d35eb6c5d2a99b95677a45053466/scalable_shampoo/pytorch}{Google Research PyTorch Shampoo, revision \texttt{5b09c22}}}
retains the single-exponent interface and does not provide distributed
execution. Meta Distributed Shampoo provides a right-only half-power update
and distributed execution in PyTorch, with the recipe differences recorded
below.

For the reported runs, each matrix statistic is initialized with
$10^{-6}I$, and each inverse root is computed after adding
$10^{-6}\widehat{\lambda}_{\max}(A)I$ to the statistic $A$. The
largest-eigenvalue estimate uses a fixed seed and a relative convergence
test. FP32 coupled Newton is used for the inverse root. A nonfinite
candidate or a final coupled-iteration error of at least $0.1$ is rejected,
leaving the preceding valid root in place. The scalar statistic is
initialized at $10^{-6}$, and its inverse fourth root follows the matrix-root
schedule. We use $\epsilon=10^{-10}$ for diagonal AdaGrad, both as the
fallback update and as the grafting donor.

Two execution paths use this backend. \emph{Replicated Scalable Shampoo}
reconstructs each full gradient and repeats the optimizer computation on
every rank. \emph{Parameter-owner Scalable Shampoo} assigns each full
parameter to one deterministic, load-balanced owner rank, where its
statistics, roots, grafting, and update are computed once. It reuses the
pinned Meta repository's FSDP2 communication contexts, including persistent
buffers and coalesced all-to-all exchanges, but retains our Scalable
Shampoo mathematics and checkpoint schema. Updated parameter shards are
returned to their FSDP2 ranks; owner-only factors are replicated only
while saving or loading a checkpoint. In a matched single-seed, 561-update
control, parameter-owner Scalable Shampoo was approximately $2.24\times$
faster in trainer time than replicated Scalable Shampoo, while validation
KL changed by less than $2\times10^{-4}$.

Meta Distributed Shampoo can also construct right-only half-power Shampoo
and its AdaGrad-grafted variant by combining per-dimension exponent
overrides with a graft donor; our Meta adapter configures these lower-level
options. They are not equivalent off-the-shelf presets for the recipes used
in the main comparison. In particular, the pinned Meta backend does not
expose our full-parameter scalar-row-mean construction. Moreover, when
blocking is enabled, Meta applies grafting per block; our Meta adapter
couples first activation to the inverse-root refresh interval; and its
default inverse-root implementation uses eigendecomposition rather than
coupled Newton. We therefore use parameter-owner Scalable Shampoo for the
main recipes and reserve Meta Distributed Shampoo as an implementation
alternative.

\begin{figure}[ht]
\centering
\includegraphics[width=0.95\linewidth]
{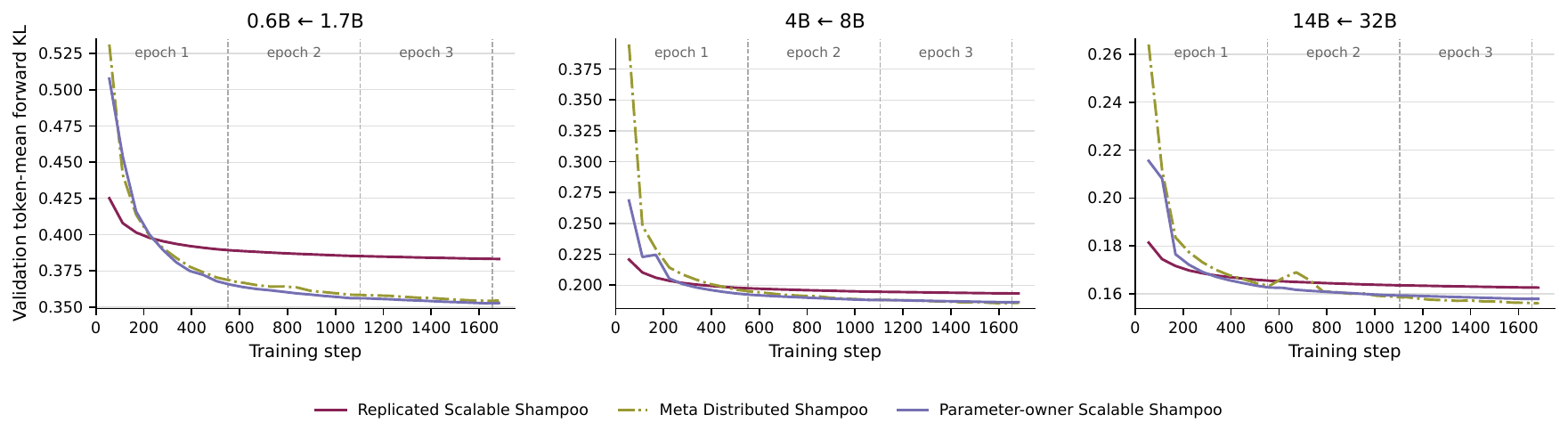}
\caption{Validation-KL trajectories for grafted right-only Shampoo.
Replicated Scalable Shampoo and parameter-owner Scalable Shampoo are
three-seed means; Meta Distributed Shampoo is a single-seed control with
additional recipe differences.}
\label{fig:shampoo-implementations}
\end{figure}

Figure~\ref{fig:shampoo-implementations} compares a grafted right-only
configuration of replicated Scalable Shampoo, parameter-owner Scalable
Shampoo, and Meta Distributed Shampoo. Parameter-owner Scalable Shampoo
and Meta Distributed Shampoo substantially reduce trainer time relative to
replicated Scalable Shampoo. Their validation behavior is recipe- and
pair-dependent rather than uniformly worse for Meta Distributed Shampoo.
Because root schedules and selected hyperparameters differ among these
historical controls, the figure documents implementation selection rather
than an isolated backend benchmark.

\subsection{SOAP}
SOAP rotates gradients and moments into Shampoo-derived eigenbases before
applying Adam along the resulting matrix directions~\cite{soap}. The main
\emph{SOAP} implementation is an unblocked FSDP2 adaptation of the authors'
implementation%
\footnote{\href{https://github.com/nikhilvyas/SOAP}{SOAP reference implementation}} and
matches it in fixed-gradient parity tests. It constructs a factor for each
matrix axis of dimension at most $8192$, retains the identity basis along
larger axes, and initializes the basis on its first optimization step
without updating the parameters.
\emph{Meta blockwise SOAP} instead uses
Meta's \texttt{DefaultSOAPConfig}%
\footnote{\href{https://github.com/facebookresearch/optimizers/tree/d24a1491797e5bf2c2985d8789f3a28293d87a6f/distributed_shampoo}{Meta DefaultSOAPConfig, revision \texttt{d24a149}}}, partitions matrix dimensions into
blocks of size at most $1024$, begins preconditioning at the first
eigenbasis refresh on update 10, and distributes the blocked optimizer
state across ranks. Its blocking and delayed use of the learned eigenbasis
change the optimizer sequence, not only its systems implementation.

\begin{figure}[ht]
\centering
\includegraphics[width=0.95\linewidth]
{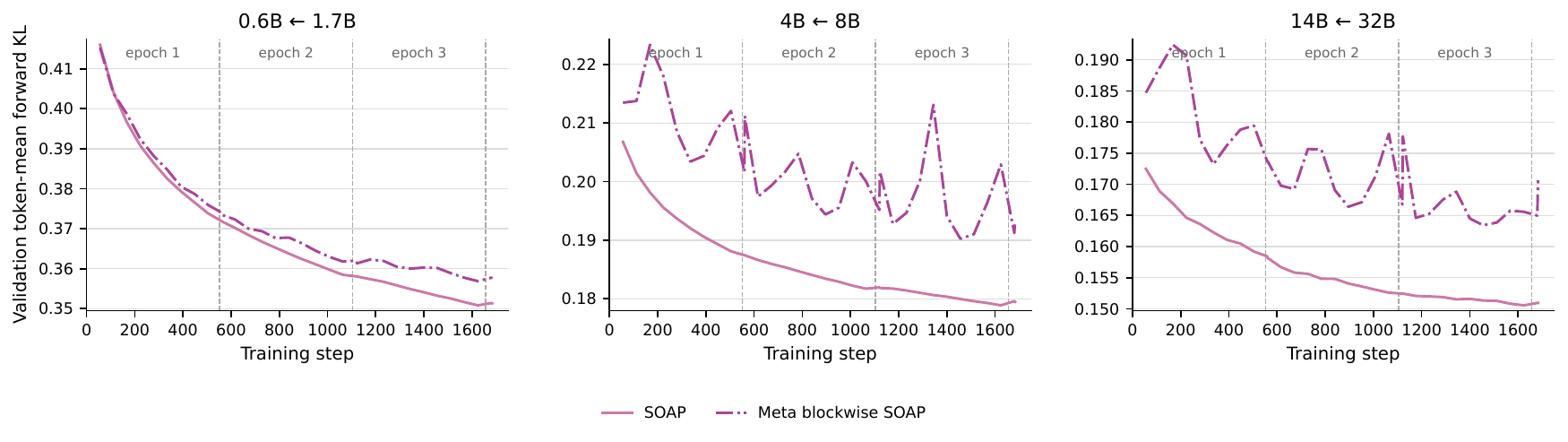}
\caption{Validation-KL trajectories for unblocked SOAP (three-seed mean) and
Meta blockwise SOAP (single-seed control). The comparison includes backend
and algorithmic differences, including Meta's matrix blocking and delayed
basis use.}
\label{fig:soap-implementations}
\end{figure}

We use the unblocked implementation because it more closely follows the
authors' implementation, retaining Meta blockwise SOAP as an alternative.
Under FSDP2, its full-matrix gradients, moments, factors, and bases are
replicated rather than partitioned.
Across the three controls in Figure~\ref{fig:soap-implementations}, Meta
blockwise SOAP achieved a $1.6$--$1.9\times$ trainer-time speedup over the
unblocked implementation but had best validation KL values
$0.006$--$0.013$ higher.
Because blocking and delayed preconditioning alter the update sequence,
this is a comparison of alternative recipes rather than an isolated
backend benchmark.

\subsection{Muon}
Muon approximately orthogonalizes matrix-valued momentum updates with
Newton--Schulz iterations. A shared distributed adapter reconstructs each
sharded two-dimensional gradient across ranks and redistributes the
update. \emph{All-2D Muon} applies Muon to embeddings, the output head, and
transformer matrices; only one-dimensional parameters use Adam.
\emph{Reference Muon} instead follows the parameter assignment of
the reference implementation%
\footnote{\href{https://github.com/KellerJordan/Muon}{Muon reference implementation}}%
~\cite{jordanmuon}: Muon is restricted to hidden transformer
matrices, while embeddings, the output head, and one-dimensional parameters
use Adam. \emph{Meta Muon} implements the same hidden-only assignment with
Meta's
\texttt{SpectralDescentPreconditionerConfig}%
\footnote{\href{https://github.com/facebookresearch/optimizers/tree/d24a1491797e5bf2c2985d8789f3a28293d87a6f/distributed_shampoo}{Meta SpectralDescentPreconditionerConfig, revision \texttt{d24a149}}} and distributes optimizer
state across ranks. The three implementations otherwise share the momentum
coefficient, Nesterov ordering, Newton--Schulz coefficients, and
aspect-ratio scaling.

\begin{figure}[ht]
\centering
\includegraphics[width=0.95\linewidth]
{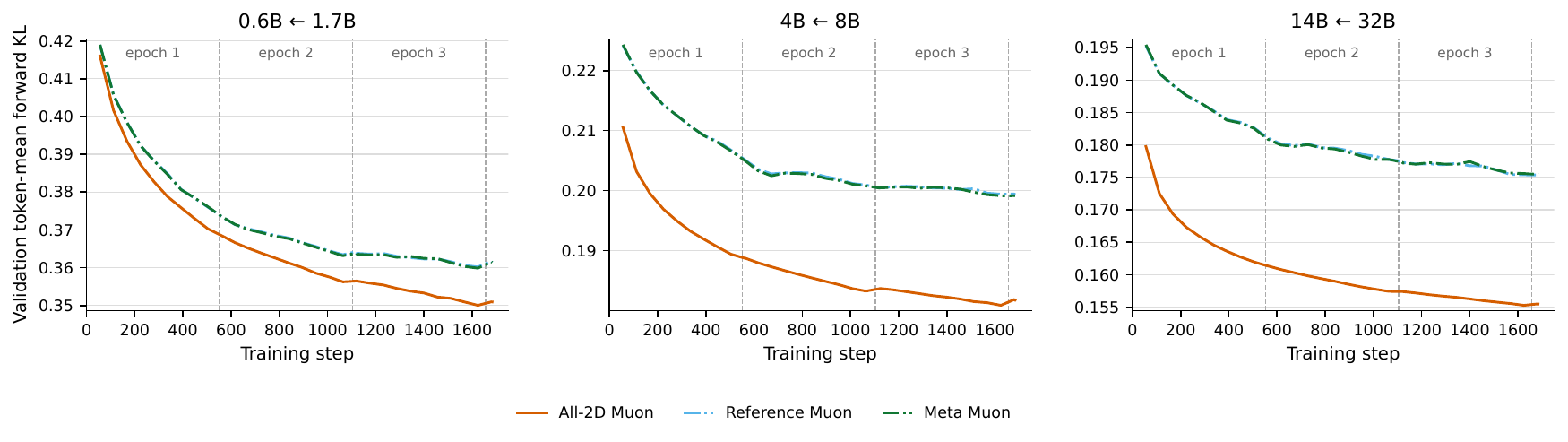}
\caption{Validation-KL trajectories for All-2D Muon (three-seed mean) and the
single-seed Reference and Meta controls. The controls apply Muon only to
hidden-layer matrices; All-2D also includes embeddings and the output head.}
\label{fig:muon-implementations}
\end{figure}

In matched single-seed controls, Meta Muon achieved a
$1.2$--$1.7\times$ trainer-time speedup over Reference Muon, while their
best validation KL values differed by at most $0.00031$
(Figure~\ref{fig:muon-implementations}). This supports agreement between
the hidden-only implementations; it does not validate All-2D Muon, whose
distinct routing means that comparisons involving its curves reflect both
recipe and implementation differences.

\section{AdaGrad-grafted right-only half-power Shampoo}
\label{app:filters}

Norm grafting takes the update direction from one optimizer and rescales it
to match the update norm of another~\cite{grafting}. Here,
$P_t=G_tR_t^{-1/2}$ is the right-only half-power direction for an eligible
matrix, and $D_t$ is the diagonal AdaGrad-preconditioned direction before
the outer learning-rate multiplier. After preconditioner activation, the
grafted variant uses
\begin{equation}
\widetilde G_t^{\mathrm{graft}}
=P_t\frac{\lVert D_t\rVert_{\mathrm F}}
{\lVert P_t\rVert_{\mathrm F}+10^{-16}}.
\label{eq:grafted-right-half}
\end{equation}
Thus grafting retains the matrix-preconditioned direction while matching
its Frobenius norm to the donor up to the stabilizer. Before activation,
eligible matrices use the donor update; parameters outside the matrix
recipe always use diagonal AdaGrad. Unlike scalar-row-mean quarter-power
grafting, this operation does not cancel the matrix preconditioner.

Grafting changes the recipe, not the backend: grafted and ungrafted
variants share the parameter-owner Scalable Shampoo execution path,
cumulative statistics,
coupled-Newton roots, and parameter routing. Grafting changes update scale,
so its learning rate and preconditioner schedule are selected independently;
the comparison is therefore between tuned recipes, not an isolated
fixed-hyperparameter grafting effect. The main comparison reports the
ungrafted variant. Completed single-seed grafted runs are exploratory and
cannot characterize seed variability or support scale-dependent conclusions.

\section{Hyperparameter selection and final configurations}
\label{app:hp}

\subsection{Selection protocol}

AdamW, AdaGrad, SOAP, and All-2D Muon were screened in short single-seed pilots
before promising settings were promoted to longer runs. Learning rates were
searched on logarithmically spaced grids spanning
$3{\times}10^{-6}$--$3{\times}10^{-4}$ for AdamW,
$3{\times}10^{-5}$--$3{\times}10^{-3}$ for AdaGrad,
$3{\times}10^{-6}$--$3{\times}10^{-4}$ for SOAP, and
$3{\times}10^{-5}$--$3{\times}10^{-3}$ for All-2D Muon, with boundary
refinement when a pilot optimum approached a grid limit. AdamW additionally
compared $\beta_1\in\{0.8,0.9,0.95\}$ and
$\beta_2\in\{0.95,0.99\}$, while SOAP compared
$\beta_1\in\{0.8,0.9,0.95\}$. Candidate configurations were ranked by the
minimum validation KL observed within their common stage budget. The initial
$0.6\mathrm{B}\leftarrow1.7\mathrm{B}$ screens used global batch 96;
final runs use global batch 144. Subsequent pair-specific pilots searched
within the same ranges. These searches select credible settings within each
recipe rather than impose an equal tuning budget across methods.

The Shampoo recipes use pair-specific staged searches over learning rate,
first preconditioner step, and refresh interval. After the 561-update
screens, the three best valid configurations are each run for 1683 updates
with three seeds. For each recipe and pair, the configuration with the
lowest three-seed mean validation KL is selected; results are not pooled
across hyperparameters.

Exploratory single-seed pilots also screened Sophia-G~\cite{sophia},
motivated by its reported pretraining efficiency, and K-FAC~\cite{kfac},
motivated by the relationship between KL and Fisher geometry. Sophia-G did
not improve the observed validation-KL--trainer-time trade-off over the
retained recipes, and K-FAC did not outperform AdamW in its control.
Neither method was promoted to the main comparison.

\subsection{Final configurations}

Table~\ref{tab:hp} records the final pair-specific values.

\begin{table}[ht]
\centering
\caption{Pair-specific learning rates and, for each Shampoo recipe, the
first preconditioner step and refresh interval (in updates; reported as
first/refresh).}
\label{tab:hp}
\scriptsize
\setlength{\tabcolsep}{3pt}
\begin{tabular}{lcccc}
\toprule
Method & \shortstack{Qwen3\\$0.6\mathrm{B}{\leftarrow}1.7\mathrm{B}$} &
\shortstack{Qwen3\\$4\mathrm{B}{\leftarrow}8\mathrm{B}$} &
\shortstack{Qwen3\\$14\mathrm{B}{\leftarrow}32\mathrm{B}$} &
\shortstack{Qwen2.5\\$32\mathrm{B}{\leftarrow}72\mathrm{B}$} \\
\midrule
\multicolumn{5}{l}{\emph{Learning rate}} \\
AdamW & $3{\times}10^{-5}$ & $3{\times}10^{-5}$ &
$3{\times}10^{-5}$ & $3{\times}10^{-5}$ \\
AdaGrad & $3{\times}10^{-4}$ & $3{\times}10^{-4}$ &
$3{\times}10^{-4}$ & $1.5{\times}10^{-4}$ \\
\shortstack[l]{Scalar-row-mean\\quarter-power Shampoo} &
$7{\times}10^{-3}$ & $5{\times}10^{-4}$ &
$5{\times}10^{-4}$ & $5{\times}10^{-4}$ \\
\shortstack[l]{Right-only\\half-power Shampoo} &
$10^{-2}$ & $3{\times}10^{-4}$ &
$3{\times}10^{-4}$ & $3{\times}10^{-4}$ \\
SOAP & $3{\times}10^{-5}$ & $3{\times}10^{-5}$ &
$10^{-4}$ & $10^{-4}$\textsuperscript{\(\dagger\)} \\
All-2D Muon & $3{\times}10^{-4}$ & $3{\times}10^{-4}$ &
$3{\times}10^{-4}$ & $5{\times}10^{-4}$ \\
\midrule
\multicolumn{5}{l}{\emph{Shampoo first step/refresh interval}} \\
\shortstack[l]{Scalar-row-mean\\quarter-power Shampoo} &
$20/200$ & $50/100$ & $20/100$ & $20/100$ \\
\shortstack[l]{Right-only\\half-power Shampoo} &
$10/10$ & $100/100$ &
$10/10$ & $50/50$ \\
\bottomrule
\end{tabular}
\parbox{\linewidth}{\vspace{2pt}\scriptsize
\textsuperscript{\(\dagger\)}Selected in the matched 561-step screen; not
extended to the three-seed, 1683-step budget because of its measured
trainer-time overhead.}
\end{table}

The remaining fixed optimizer-specific settings are:
\begin{itemize}
  \item \textbf{AdamW:}
  $(\beta_1,\beta_2)=(0.8,0.99)$ and $\epsilon=10^{-8}$.
  \item \textbf{AdaGrad:}
  $\epsilon=10^{-10}$ and zero initial accumulator.
  Unlisted AdamW and AdaGrad arguments use PyTorch defaults.
  \item \textbf{Both Shampoo recipes:}
  cumulative statistics updated every step, no momentum or grafting,
  maximum matrix-factor dimension $8192$, matrix-statistic initialization
  $10^{-6}$, and diagonal-fallback $\epsilon=10^{-10}$. Scalar-row-mean
  quarter-power Shampoo also initializes its scalar statistic to $10^{-6}$.
  \item \textbf{SOAP:}
  $(\beta_1,\beta_2)=(0.8,0.95)$, Adam-denominator
  $\epsilon=10^{-6}$, factor EMA coefficient $0.95$, and eigenbasis refresh
  every 10 updates.
  \item \textbf{All-2D Muon:}
  momentum $0.95$ with Nesterov acceleration, five bfloat16 Newton--Schulz
  iterations, and one-dimensional Adam with
  $(\beta_1,\beta_2)=(0.9,0.95)$ and $\epsilon=10^{-10}$.
\end{itemize}

\end{document}